# Generative AI translations in high-stakes emergency messaging

Nune Ayvazyan (Universitat Rovira i Virgili)
Yu Hao (University of Melbourne)
Anthony Pym (Universitat Rovira i Virgili)

Abstract

Emergency messaging such as extreme-weather reports and earthquake instructions can involve high stakes, to the extent that translation errors can lead to tragic consequences. The use of machine translation or generative artificial intelligence might therefore not be recommended. On the other hand, time savings in the initial translation can allow greater investments of resources in revision and authorization processes, as well as a wider range of target languages. An experiment with generative AI translations of an earthquake instruction text from English into Chinese and Spanish shows that use of discourse-specific prompts can considerably improve understandability and actionability, although the translations may still not be trusted by translators. Human revision is still required, not only to detect errors but also because of the ethical need for someone to take responsibility for any errors or delays in such messaging.

## Introduction

On March 1, 2025, a Presidential Action declared English to be the official language of the United States (White House, 2025). Just over one month later, on April 9, 2025, the United States National Weather Service (NWS) stopped translating emergency warnings into languages other than English (CNN, 2025). That might have been a practical application of the one-language presidential policy, perhaps operating in conjunction with federal government cost-saving, although the official reason we received was simply that there had been a "pause" in the service "due to a contract lapse" (Musher, 2025). This is of interest because the United States National Weather Service was offering what it described as "real-time" translation by generative artificial intelligence into five languages, through a website (weather.gov/translate) where the user selected the desired language and the "product", choosing from a menu that included hazard maps and alert banners for risks in the user's selected area.

How did that system work with respect to the translation process? Buchanan (2023) and Yarborough (2023) describe the AI company Lilt spending two years compiling databases in the two main target languages: Spanish and Chinese. An AI system then drew on the databases to produce translations. In so doing, it replaced a workflow based on unaided human translations, which were considered labor-intensive, slow, and "prone to error" – which implies that the AI translations were less prone to error. Buchanan (2023) offers a description of the system that suggests the AI translations themselves were in fact not all produced in "real time", if one understands that term to imply the use of raw automated output. Instead, it allowed a reallocation of time-on-task: “our Spanish forecasters reduced the time needed to translate National Hurricane Center storm products from 1 hour to less than 10 minutes, which allows them to focus more time on their core duties of forecasting and decision support services”. This means that not only were there were serious time savings, but at least some of the translations were being checked by humans – since something was done in those ten minutes. The description of the process then becomes a little confusing. On the one hand, the new AI system allowed “forecasters to more easily check and edit the translations”, which sounds fine; on the other, it provided “translations in languages for which NWS does not currently have bilingual staff and without hiring full-time translators”. This suggests that at least some of the output was either not checked (hence the mention of “real time translations”) or, more ideally, external translators were employed to revise and authorize the translations, at speed.

This type of emergency messaging is of interest because weather forecasting was perhaps the first clearly successful application of machine translation: the Canadian French-English weather reports system Météo had rule-based machine translation operating from May 1977 and remaining in use through to 2001. An early report (Thouin, 1982, p. 44) described it as rendering about 80% of the information and sending the rest to human translators or revisers. Weather reports would seem almost ideal for machine translation: a fixed, well-defined terminology with established equivalents in many languages, restricted tenses and verb aspects and, in the case of Météo, telegrammatic syntax. Beyond those linguistic criteria, meteorological messaging can be very high-stakes (in the case of hurricane warnings and the like), its events are repeated (terminology and style sheets can be developed over time), and the messages themselves have a very short shelf-life (about six hours in the case of Météo – the translations were only actionable if they were published within that time frame). Few domains would seem so well suited for automated translation.

Here we ask how far that set of advantages can be extended into a wider set of emergency messaging, and how far the ostensibly successful use of machine translation can be extended into the use of generative artificial intelligence. We report on an experiment we carried out with a text that gives instructions on how to survive an earthquake – similar to weather forecasting in some respects, and with perhaps even higher stakes if the automation gets things wrong. We ask what errors are made with generative AI when translating from English into Chinese and Spanish, what errors can be mitigated through the use of AI-generated prompts, how time and costs can thus be reduced, and especially what the role of humans should be in a highly automated high-stakes workflow.

**Previous studies**

The experiment we are reporting on here emerged from our previous research on healthcare messaging in the COVID context of Melbourne (Hajek et al., 2022; Karidakis et al., 2022), where more than 270 languages are spoken in homes. Data from over fifty interviews with community representatives indicated the limitations of unaided human translations in time-constrained situations involving high numbers of languages; the interviews also made us acutely aware that translations need to be trusted if they are to be a factor in behavior change (Pym et al., 2023). In the healthcare context, we detected a policy move towards the use of community-based mediators, some of whom would pick up machine-translated messaging and then reformulate it in ways that were more accessible to their community members, often with a switch from written to audiovisual media (Pym and Hu, 2022).

Following that project, we looked more specifically at the use of machine translation in emergency messaging. Using community-based focus groups and language informants, we tested the understandability and actionability of machine-translated emergency messaging (warnings of bushfires, sharks, and water contamination) from English into Chinese, Greek, Arabic, and Dari (Hajek et al., 2023). In parallel, an independent project in Spain carried out a linguistic evaluation of unrevised Google Translate output in healthcare messaging from Catalan into Spanish, English, and French (Pym, Ayvazyan, and Prioleau, 2022).

Both those later projects demonstrated the viability of *pre-editing* as a way of working with neural machine translation, showing how original texts can be written in such a way that automated translation works better into a range of very different languages at the same time. While the main recommendation of those projects was that original texts be written in accordance with a set of basic rules concerning linguistic clarity and text layout (Sengupta et al., 2024), we also explored several other ways in which machine translation could be integrated into time-sensitive workflows. Emergencies are always unexpected, but they are repeated. This means it is relatively easy to develop and translate *pre-written templates* that give the information chunks that are then mixed and matched in any particular message, with specific details being filled in for the new event. It made sense to translate those templates into the main target languages at potential risk well before there was any immediate crisis. The workflows could then include a stage of *human post-editing and authorization*, as well as systems for community-based relaying and explanation of information verbally, as already exists in the case of bushfire messaging in Australia. These different kinds of workflows, integrating automation and human translation and/or revision in various ways, were each seen as offering a specific advantage. This meant that actual workflows can be selected in accordance with the risks and the resources (especially time) available in each specific case, applying some of the principles of triage used in medical situations (Pym, 2023).

The extension of this research into the use of generative AI is in response to new technological capacities. Our previous study on machine translation (Hajek et al., 2024) showed that generative AI could solve some problems that neural machine translation struggled with, especially the inconsistent use of the formal and informal second persons in Chinese and Spanish. Further, our work with focus groups had indicated instances where errors made by neural machine translation could have led to fatalities, and two of those major errors were also made in generative AI translations – although they were no longer in evidence in output from the same systems three months later. At that time, research by Hendy et al. (2023) and Jiao et al. (2023) was showing that some versions of generative AI could score better than some versions of neural machine translation for some kinds of texts, but that this advantage only held for languages with extensive electronic resources.

We were therefore interested in testing how well generative AI could perform as an effective way of translating high-stakes emergency messaging. Could we perhaps identify something more cost-effective than the generative AI system that took two years, with a corresponding budget, to set up for the United States National Weather Service?

## Methodology

We selected an 892-word text that gives instructions on what to do in an earthquake. The English-language original was downloaded in October 2024 from the website of the Victorian government in Australia (ses.vic.gov.au/plan-and-stay-safe/emergencies/earthquake). The text was a set of standard long-standing instructions, not subject to any particular shelf life, although its translations into further languages might conceivably be undertaken under extreme time constraints if and when a particular language community is affected by an earthquake. Remarkably, the government website carried no official translations of this text, although a drop-down menu offered *general* emergency information in 40 languages, including the number of an emergency hotline.

We then ran that English-language text through ChatGPT-4o with the default prompts *Translate this text into Chinese* and *Translate this text into Spanish*. We selected the ostensibly erroneous, ambiguous, or otherwise unhappy renditions in those translations and included them in a reduction of English text to 442 words (Appendix A). The raw translations into Chinese and Spanish were reduced accordingly (Appendices B and C), ensuring that they rendered the reduced text of 442 words. In doing this, we consciously made the translations appear globally worse than they were, since we gave them higher densities of what we thought could be errors. We aimed to produce translations that could provide information on the reception of problematic items without the informants losing concentration. We were aiming for an experiment session of about 40 minutes.

Those reduced translations (Appendices B and C) were then shown, *without* the English-language original, to 62 students in the Master of Translation and Interpreting program at the University of Melbourne, Australia, all of whom were L1 speakers of Chinese, and 55 third-year undergraduate students at Universitat Rovira i Virgili in Tarragona, Spain, all of whom had L1 competence in Spanish.[1] The students were asked a series of multiple-choice questions testing the understandability and actionability of the translations, which they were requested to respond to from the perspective of end-users (i.e., thinking about how to survive an earthquake). This gave us basic points of comparison for evaluating the generative AI translations in the two languages.

The participants were then shown the reduced English original (Appendix A), which they used to revise the generative AI translations manually. At this stage, they were also told how those translations had been produced.

In subsequent steps, we used a basic agentic model to have ChatGPT revise its previous translations. We did this by using two prompts: *List the discourse and stylistic features of the best warning texts in simplified Chinese for PR China / Spanish for Spain,* and then: *Use those features to revise the previous translation*. That is, we asked it to generate a detailed discourse-specific prompt. This gave us two AI-revised translations (Appendices D and E). The participants were invited to compare their own revisions with the AI revisions, after which we led a general discussion on the issues involved.

That activity was designed as a pedagogical exercise to test to what extent students could learn from using generative AI for revision, an aspect that has been reported on elsewhere (Raigal et al., 2026; Hao and Pym, forthcoming). In the present context, it also gave information on the extent to which the students were prepared to trust the revisions proposed by generative AI. That is, we were able to gain data not only on the understandability and actionability of the generative AI translations, but also on their perceived trustworthiness.

It is important to note that all our experiments, including the one being reported on here, have evaluated the quality of translations on the basis of feedback from human participants, in the form of both quantitative answers to surveys and qualitative responses in interviews. More traditional studies use automatic or semi-automatic measures such as BLEU, METEOR, or TER scores, which we consider problematic because they rely on a reference human translation – when, in most cases, human translations can be different yet equally viable, in addition to the errors that all humans make. Some methods are based on how much work is done by post-editors (following Krings, 2001), which is also problematic because of the variability of post-editing criteria. Instead, here we focus on what really counts for this kind of text: can the receiver understand it, trust it, and therefore act upon it?

[1] Ethics clearance was received from the Human Research Ethics Committees, Faculty of Arts, University of Melbourne (2024-27953-49908-4) and the Ethics Committee for Research on People, Society, and the Environment at Universitat Rovira i Virgili (CEIPSA-2024-PR-0012).

## Results

As mentioned, the first evaluation was carried out by having our participants read the ChatGPT translations into Chinese and Spanish, *without seeing the original English text*. They then answered six multiple-choice questions concerning the actionability of problematic phrases in the translations. Two of those questions were virtually the same in both the Chinese and English, while the remaining four were specific to one language or the other.

*How does a body drop?*

Our first question concerned the first core instruction, which was translated as follows:

> ST: Drop to the ground.
> TTzh: 趴下 [pa-xia, get down]
> TTes: Agáchate al suelo. [crouch/squat to the ground]

We asked the students: "Which of these images does *not* depict the translation. Choose as many as applicable":

| A. | B. | C. | D. | E. Not sure |
|---|---|---|---|---|
| 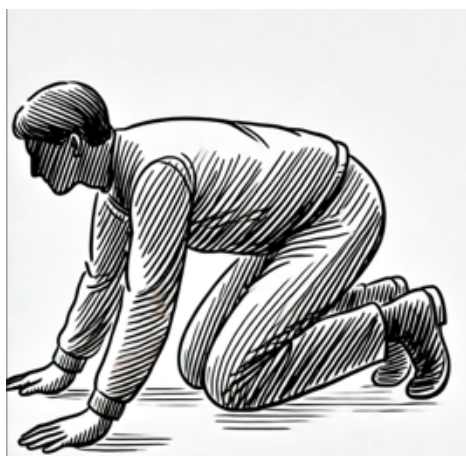 | 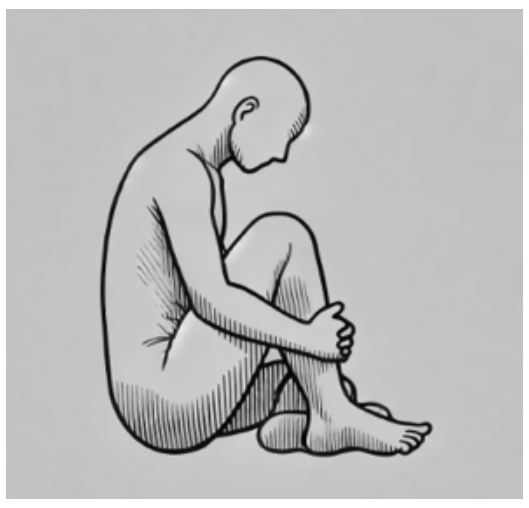 | 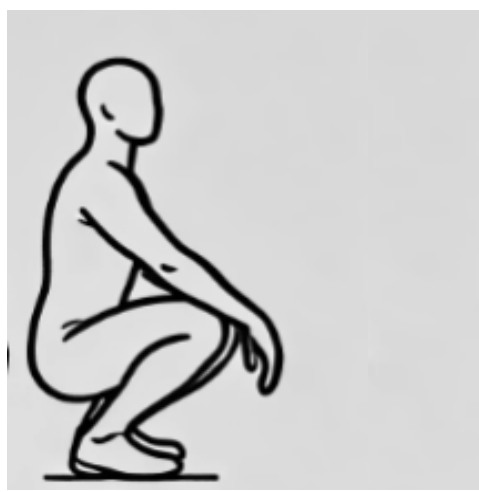 | 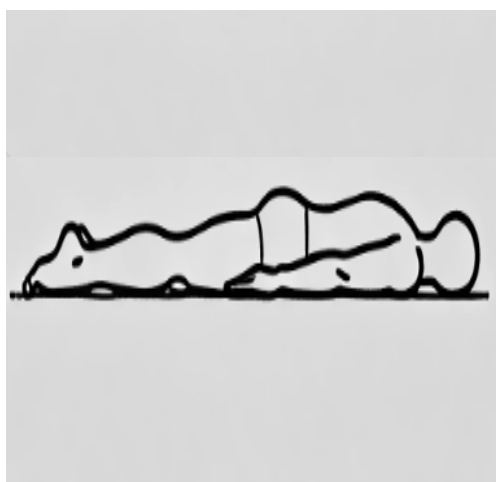 | |

The problem here is that the Chinese translation "趴下" refers to A and D (the latter would be particularly risky in an earthquake) but not to B and C. The English-language text makes it clear that A is good because it is the position you need in order to crawl under a table, and B is good if you are against a strong wall. That is, the Chinese translation covers one correct possibility and one position that could be dangerous, especially if you are left looking downwards. Meanwhile, the Spanish translation literally means "crouch/squat to the ground" and could apply rather vaguely to A, B or C, all of which would keep one reasonably safe.

The participants were asked to choose as many options as applicable for this question, yielding a total of 144 responses for Chinese and 76 for Spanish. Only four of the Spanish responses included the dangerous position D as a possibility, so the translation might be considered generally actionable. Almost half the Chinese responses (48%, N = 69), however, included dangerous D, indicating that they erroneously believed the position could help them safe. Further, 52 of the 62 Chinese respondents did *not* opt for option B (the "crouching" body gesture), indicating a clear mismatch between the original and the default genAI translation.

*What is a doorframe for?*

An important instruction in the English text had the following default genAI translations:

ST: Do not use a doorway except if you know it is a strongly supported, load-bearing doorway and it is close to you.
TTzh: 8%
TTes: No uses una puerta a menos que sepas que es una puerta con soporte fuerte, de carga, y esté cerca de ti.

The Spanish has a clear error: *puerta* basically means "door", rarely "doorway", although the missing specificity could be recuperated from the rest of the sentence. When asked to choose between three possible actions, the participants interpreted the translation of *Do not use a doorway* as: 1) don't open a door (19%), 2) don't go through a doorway (27%), 3) don't shelter under a door frame (44%), which was the correct answer, and 4) not sure (5%). That is, only 44% interpreted the instruction correctly, making this a high-frequency and high-stakes error.

Similarly, the default Chinese translation omits the required reference to a frame and, even worse, introduces an uncommon verb-object collocation. The interpretations were: "don't take shelter under the door frame" (66%), "don't stand at the entrance" (25%), and "don't stand at the front door" (8%). That is, a third of the participants would not have taken the correct action.

*Hold on to what?*

The main instruction repeats three verbs: *drop*, *cover*, and *hold on* (harking back to the slogan "duck and cover", which Americans once hoped would keep them safe after a nuclear attack). In repeating the slogan, though, the object of the verb becomes implicit – the receiver must retrieve it from the co-text.

ST: Drop, cover (under a sturdy table or piece of furniture) and hold on.
TTzh: 趴下；找一个坚固的桌子或其他家具进行掩护；并抓紧，直到震动停止。
TTes: Agáchate, cúbrete (bajo una mesa resistente o un mueble) y sujétate.

We asked the Spanish speakers whether *sujétate* (hold on) means: 1) hold on to yourself (since the verb is reflexive), 2) hold on to a fixed object, 3) cover your head with your hands, 4) not sure. More than three-quarters of them (38; 79%) correctly opted for "hold onto a fixed object" (since furniture has just been mentioned), but that means that 21% did not construe the phrase correctly.

*Cover what?*

In this same sentence, *cover* was rendered as "掩护" (yan-hu) in Chinese. The participants were asked whether this meant: 1) to provide cover fire to allow comrades to move safely; 2) cover up or support someone; 3) find an object or structure that can shield or protect you; and 4) not sure. The vast majority (61; 94%) went for the right answer: find an object or structure that can shield or protect you. This is despite the fact that the Chinese verb tends to be used as a war or wargaming term, meaning to provide cover fire. That is, even though the wording leaves something to be desired, the participants could still correctly interpret the intended message from the context.

*Where do you proceed cautiously?*

The apparently innocuous phrase "proceed cautiously" proved problematic in Spanish but not in Chinese:

> ST: If you are in a moving vehicle during an earthquake: Stop as quickly as safety permits […]. *Proceed cautiously* once the earthquake has stopped.
> TTes: Si estás en un vehículo en movimiento durante un terremoto: Detente tan rápido como sea seguro […]. *Continúa con precaución* una vez que el terremoto haya terminado.

Our Spanish-speaking respondents interpreted the phrase as: 1) stay where you have stopped (0); 2) drive further on in the same direction (12; 25%); 3) keep being cautious (35; 73%); and 4) not sure (1; 2%). In the English text, the car has stopped and then starts moving again, *cautiously*, when there is no more shaking. For about three quarters of our respondents, however, the Spanish text said the car should remain stationary, and that the caution should continue. There is a clear problem of actionability here. At the same time, since erroneous interpretation would lead to non-action, there is no major risk involved.

*Why a simple list?*

People with what are termed "developmental/cognitive/intellectual disabilities" receive the following instruction in English and Chinese (the Spanish is fine):

> ST: If you have difficulty understanding, remembering, or learning, keep a simple list of what to do.
> TTzh: 请随身携带一份简单的待办事项清单。

The participants were asked if this means: 1) you should include a workplace to-do list in your emergency kit; 2) you should carry a guide for action in your kit; 3) you should carry a list of pending matters during the earthquake; or 4) not sure. In this case, fewer than half the participants (30; 46%) correctly inferred that the list should be a guide that cognitively impaired people could use to remember what actions should be taken when an earthquake strikes. The majority were unable to construe that meaning from the sentence.

Given these responses, how well did the default translations fare? The numbers in bold in Table 1 indicate serious problems with respect to the participants' construal actions to take, although high levels of risk are only approached in the cases of "drop to the floor" and "do not use a doorway".

Table 1. Actionability answers of 65 Chinese-speaking and 55 Spanish-speaking participants

| | **Chinese** | | | **Spanish** | | |
|---|---|---|---|---|---|---|
| | Correct | Not sure | Wrong | Correct | Not sure | Wrong |
| "drop to the floor" (multiple selections) | 69 (48%) | 1 (%) | **75 (51%)** | 69 (84%) | 4 (8%) | 4 (8%) |
| "do not use a doorway" | 43 (66%) | 0 | **22 (34%)** | 21 (44%) | 5 (10%) | **21 (46%)** |
| "hold on" | | | | 38 (79%) | 4 (8%) | 6 (12%) |
| "proceed cautiously" | | | | 12 (25%) | 1 (2%) | **35 (73%)** |
| "cover" | 61 (94%) | 0 | 4 (6%) | | | |
| "a simple list of what to do" | 30 (46%) | 0 | **35 (54%)** | | | |

In short, although many readers can successfully make sense of flawed translations, at least two errors remained problematic and could be considered life-endangering.

**Can discourse-specific prompts ensure comprehension and actionability?**

As mentioned, there are several ways of improving the content of generative AI translations. Our previous work mainly explored pre-editing, while other researchers have invested serious efforts in the use of curated training data, as in the case of the National Weather Service (Buchanan, 2023) and as proposed for the weather forecasting system in Canada (van Beurden, 2019). Here we were looking for a cheap, fast, and efficient way of improving the translation by using the resources in the technology itself, rather than resorting to the corpus-based methods that could belong to a paradigm of the past.

As mentioned, we asked GPT to list the best practices for emergency messaging in the two languages and then to use those features to revise its previous translations. Would that be enough to produce some viable translations?

Yes and no. The retranslations made several laudable stylistic changes, using more bullet points, block language, and simpler sentences. In Spanish, the headers of each section were converted from conditionals (e.g., back-translation: "If you are indoors") into questions (back-translation: "What to do if you are indoors?"); in Chinese, the main points were presented in bold and the action instructions were moved to the beginning of paragraphs and sentences, as is expected in Chinese. The main problems were nevertheless semantic, not stylistic. Here we survey what happened to the items we have looked at above.

The translation of *drop to the floor* remained unchanged in Chinese and was not effectively resolved by the slight variant introduced in the Spanish retranslation: *agáchate en el suelo* ("crouch/squat *on* the ground") instead of *agáchate al suelo* ("crouch/squat *to* the ground"). Neither of these is abundantly clear (there were many opinions in the discussion session), but both are at least sending you in the right direction, and they will not leave you lying face-down. A potential solution to the semantic mismatch in Chinese would be to amplify the one English verb as two Chinese verbs: "趴下" (the literal translation) plus "蜷缩" (huddled up, holding on to yourself). This would include option B. Alternatively, to solve the problems in both languages, the English could have been written as *sit or crawl on the floor*, albeit at the cost of losing the memorable slogan comprising the blunt, vague verbs *drop, cover, hold on*. A neat trade-off would be to use *drop* the first time for the slogan, then elaborate with *sit* (when there is no table) and *crawl* (when there is a table). In sum, much more thought needs to be put into how to write the original text with a view to subsequent translations.

A trickier case is the generative AI translation of *Don't use a doorway* as "不要使用门口", which was wrongly interpreted as "don't stand at the entrance" (25%) or "don't stand at the front door" (8%). The revised generative AI version was "不要站在门口" (don't stand in the doorway/at the door), which was clearer but still by no means unambiguous. To be interpreted correctly, it needs to be read in relation to the concept of "load-bearing", in which case the reader understands not to take shelter under the door frame because it may be weak.

Everything else, however, was either solved in the retranslation, more or less, or was of minor consequence (in the sense that there were low percentages of wrong answers). The *puerta* that was seen as a door rather than a "doorway" was correctly rendered as *marco de puerta*, the "frame of a door". The problematic *continua con precaución* was not changed (although ChatGPT shame-facedly claimed it had changed it to *procede con precaución*), but that was a low-stakes problem anyway. The Chinese for *cover* also remained unchanged, which is not too bad given that it already had a 94% actionability score. And the *simple list* became 简单的安全步骤清单 (a simple checklist of safety steps), where the addition of 安全步骤 (safety steps) definitely helps with the disambiguation.

In sum, the discourse-specific prompt gave a translation that is a marked improvement with respect to both understandability and actionability, although it is still by no means perfect.

In this kind of text, though, 'almost' is not good enough. Not only is there at least one remaining infelicity that could result in a wrong action in at least a small percentage of cases, but there are quite a few instances in which actionability could be enhanced by rewriting the original English and/or adding explications in the translations. That is, understandability and actionability still require human intervention in the workflow.

**Are genAI translations trusted?**

As mentioned, our previous work on the social reception of translations looked not just at actionability and understandability, but also at the ways in which some messaging was trusted or distrusted. We found that, in some communities, a translation could be understandable and actionable but not trusted and hence failed to contribute to behavior change. The reasons for distrust were culturally variable (type of language, identity of the sender, identity of the mediator), but they tended not to include the automation of translation – there are many situations where minority communities choose to rely on machine translation rather than on an unknown interpreter (Pym, 2025).

In the present case, we can make some inferences about trust from the ways our participants compared their own revisions with the retranslations made by genAI, as well as from their general assessments of how well the technology performed. We offer a few salient examples of what we found to be a general tendency.

As noted, the default AI translation of *doorway* as *puerta* in Spanish led most of the participants to opt for the wrong action. This could explain why only 13% of them revised that translation – they probably did not see the problem. Those that did revise it reported that the AI revision was justified and "similar" or "better" than what they did. Of those who did not revise the phrase, 46% admitted "it is something I should have thought about", which is good news for pedagogy – human translators can learn from AI translations. However, a full 30% of that group found the AI revision *unjustified*, selecting between drop-down options to indicate that it "makes no sense" (13%), it "unjustifiably adds something" (10%), it "makes no difference" (10%) or it "is misleading or wrong" (6%). That result, puzzling linguistically, suggests that the automatic output reached a point here where it was simply not being trusted. This might have been because the participants knew it had come from a machine.

In generative AI retranslation, the phrase *drop to the ground* was made clearer in Chinese but not in Spanish. Only 31% of the Spanish participants revised this phrase, which implies that more than two thirds saw no problem with it, and yet when selecting from the drop-down menu of comments on the AI revised translation, those who had made no change were again remarkably negative: 29% said "it is misleading or wrong"; 25% opted for "it makes no sense"; 23% said "it makes no difference"; while only 10% said "it is something I should have thought about". In this case, there would appear to be even less trust, and considerably less learning – although neither of the translations was particularly clear.

Rather different things happened on the Chinese side. The generative AI revision of the doorway instruction as "不要站在门口" (don't stand in the doorway/at the door) was scarcely satisfactory in terms of understandability, but 44% of the participants nevertheless believed it was better or much better than their revision, while 27% stated it was as good as their revision. Why so much positive appreciation? Choosing between the drop-down options, 25% of the participants said the generative AI revision was "more idiomatic" than theirs; 22% thought it was "more accurate"; 16% said it was "something I should have thought about"; only 5% opted for it being "misleading or wrong". In this case, we suspect the results indicate an *excess* of trust. The generative AI revision was stylistically more natural, and this fluency seems to have masked considerable semantic imprecision.

As these examples suggest, we generally found that the Spanish participants were less disposed to trust the generative AI revision than were the Chinese. This difference comes out in the responses to the question "How well do you think ChatGPT revises translations?", which we asked both groups before and after the experiment. The results (Figure 1) show that the Chinese speakers started from a general expectation that the technology would perform *better* than them, whereas the Spanish speakers initially tended to think the technology would perform *worse* than them. The experiment moved the assessments of the two groups in opposite directions.

Figure 1. How well do you think ChatGPT revises translations?

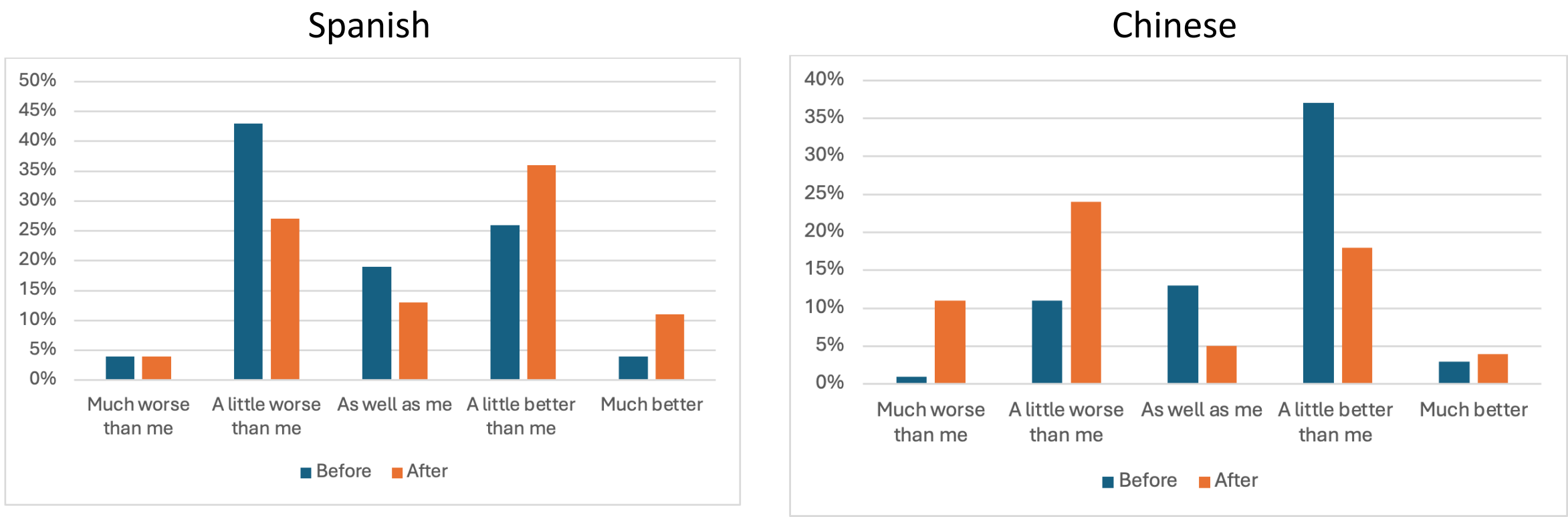

What could explain the opposed directions of these movements? It could have been due to differences between the languages and, thus, differences in the quality of the translations. And yet, when we asked the similarly basic question of the extent to which generative AI should play a role in the translation of this kind of messaging, there was no significant difference between the two language groups (Figure 2). We are left to surmise that the fundamental difference lies in the variable cultural inclinations to trust automated translations.

Figure 2: To what extent do you agree that ChatGPT should play a role in the revising of the translated emergency messaging?

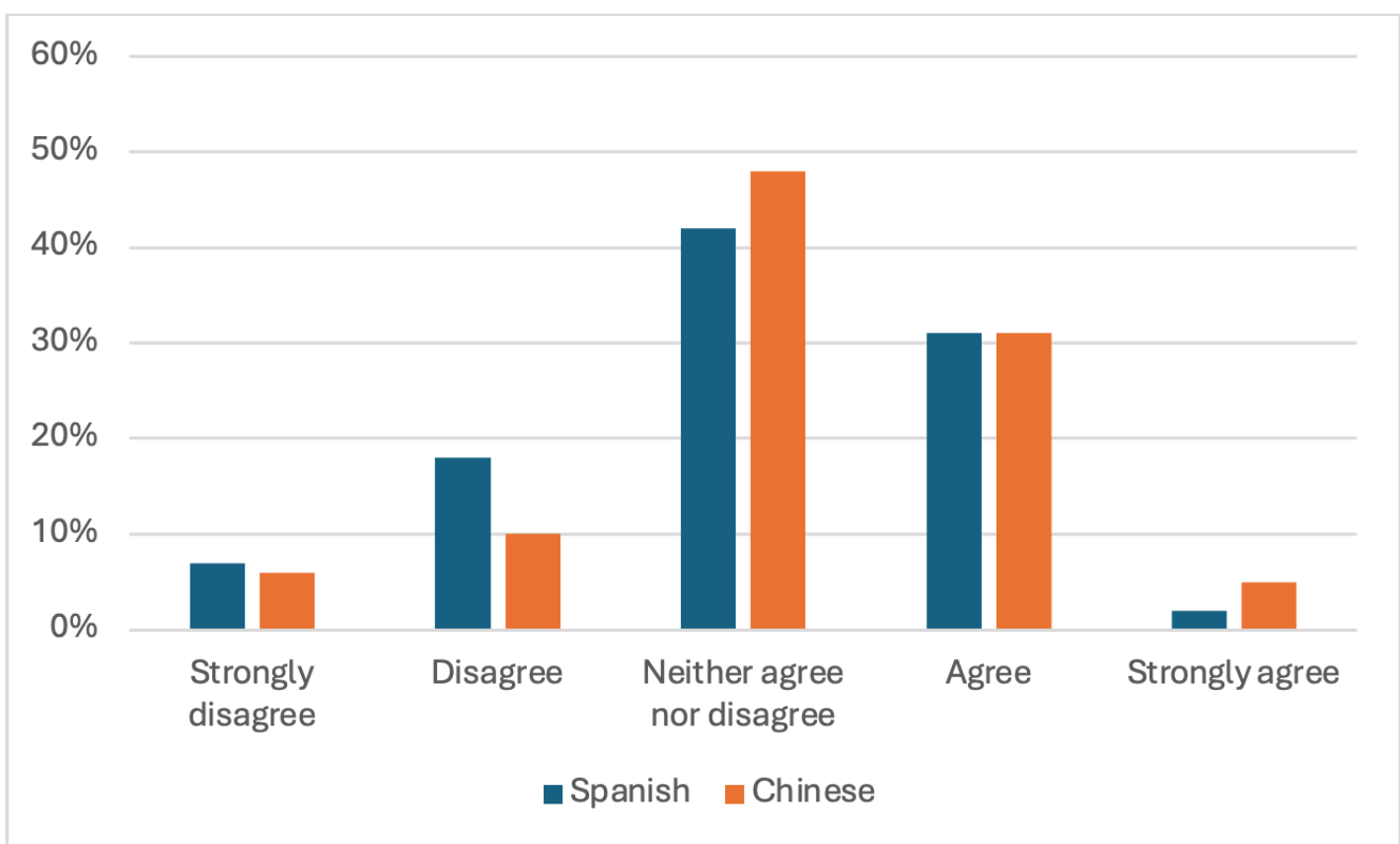

In the discussions that followed the experiments with these texts, a range of opinions were expressed. As indicated in Table 2, there was a group in both languages that was clearly against automation. A Chinese participant commented that "a fully human revision is always the best”; a Spanish participant echoed this: “I think a human should do it”; another Spanish speaker was keenly aware of the risks involved:

> I rate ChatGPT revision as horrible. Whenever you’re in an emergency setting, decisions are important in the way they’re made, and with the flux that ChatGPT has people could’ve lost their lives. That shouldn’t happen at all. The things with the door, or laying on the ground. Someone could be dead.

That, however, was a minority position. Most participants argued that generative AI could be used, but in combination with human revision at some stage. There were various justifications for this. One Chinese participant emphasized the urgency factor: “Time matters a lot here. GenAI can revise the existing translation more quickly than any human revisors.” A Spanish speaker sought a more general balancing act:

> As long as it has human supervision. There were some things, like the example with “agáchate”, the door thing… there are some imprecise directions that could threaten people’s lives. There are important things to be fixed.

This general trade-off position combined the advantages of automated revision and human revision. Given the initial differences in expectations of generative AI, the basic trade-off could be seen as a point towards which both groups moved, albeit from different directions. We might describe that point as informed low or vigilant trust – not distrust, since there is no suggestion of anyone being betrayed, but it is not fully blind trust either.

There were, however, marked differences in the discussions. Chinese translation classes are usually non-political, since there are many sensitivities involved when students come from different parts of the Chinese-speaking world. Not so in Spain, however. There, the discussion turned to the problem of accountability, with references to failed warnings of fatal flash floods in Valencia in 2024, for which no one had been declared responsible, and our discussion actually occurred two days after a major power outage in Spain and Portugal, for which no one had been identified as being at fault. That was the context in which one participant gained general agreement for the following:

> There should be accountability. For example, what happened in Valencia, people received the messages late and the messages were not specific enough about what was happening, so now there are judicial processes against those people who wrote those messages and sent them out, because they are guilty of that. If it was only ChatGPT doing it, nobody is really at fault besides whoever wrote the prompt. This means people who write these messages must have some stakes in it, because they have it in their best interest to do things correctly.

The point seems well made. The more automation we have in our communication, the harder it is to pinpoint responsibility. Indeed, the mechanical anonymity that provides one reason why some groups place trust in machine translation (because they distrust human mediators) can, in other circumstances, become a smokescreen for the evasion of ethical responsibility.

## Conclusion

When Météo was machine-translated in Canada in the 1970s, it took two years to set up the system and get it running to the point where it could automatically translate about 80% of the input (Thouin, 1982). When the United States National Weather Service was automated using generative artificial intelligence, it also took two years to set up and have running in Chinese and Spanish (Buchanan, 2023). Those two years represent considerable costs. That is why we have experimented here with a translation workflow that effectively uses the investments already made in a generative AI system. We have shown how that system can improve translations in a way that could respond to urgent needs. More specifically, generative AI can be a part of workflows that are fast, relatively cheap, able to attain reasonable levels of understandability and actionability when used with human intervention, and can be socially functional when approached with a regime of low, vigilant trust.

That said, if we were to spend two years developing this kind of system, most of our efforts would be put into standardizing the way technical writers *draft* texts for automated translations (Hajek et al., 2024). With fewer slogans and some much clearer explanations in the original English, many of the above problems could be solved before they appear.

Our experiment has several clear shortcomings. We compromised the authenticity of the original text by concentrating the problematic passages for the sake of the experiment; we have worked with the world's three major languages, where electronic resources abound and generative AI can be expected to perform better than with other languages; we have relied on university-level bilingual students as informants in matters of quality, whereas actual receivers would have far more diverse profiles; and we have incorporated their perspective as non-professional human translators as a measure of understandability, actionability, and trust, on the assumption that acceptance by translators will be needed before the technology can be integrated successfully.

A further problem is that our translations appear to be time-bound. Some of the GPT errors that we identified in October 2024 were no longer present in April 2025 – the 'dropping' problem remained, for example, but *door* had clearly become *frame of a door*, even without a discourse-specific prompt. This is in line with reports of rapid progress in the technology, although we also suspect that, as in quantum mechanics, the act of observation affects the object observed.

*Appendix A: Original text reduced to 442 words*

**If you are indoors during an earthquake**
- DROP to the ground; take COVER by getting under a sturdy table or other piece of furniture; and HOLD ON until the shaking stops.

Stay indoors until the shaking stops and you are sure it is safe to exit.
Avoid exterior walls, windows, hanging objects, mirrors, tall furniture, large appliances, and kitchen cabinets with heavy objects or glass.
Do not use a doorway except if you know it is a strongly supported, load-bearing doorway and it is close to you. Many inside doorways are lightly constructed and do not offer protection.
Stay inside until the shaking stops and it is safe to go outside. Do not exit a building during the shaking. Research has shown that most injuries occur when people inside buildings attempt to move to a different location inside the building or try to leave.
Be aware that the electricity may go out or the sprinkler systems or fire alarms may turn on.
**People with Developmental/Cognitive/Intellectual Disabilities**:
If you have difficulty understanding, remembering, or learning, keep a simple list of what to do and important information with you and in your kits. Practice your plan in advance. If you use augmentative communication supports, include these in your planning.
**If you are outdoors during an earthquake:**
Stay there.
Move away from buildings, streetlights, and utility wires.
Once in the open, stay there until the shaking stops. The greatest danger exists directly outside buildings, at exits and alongside exterior walls.
**If you are in a moving vehicle during an earthquake:**
Stop as quickly as safety permits and stay in the vehicle. Avoid stopping near or under buildings, trees, overpasses, and utility wires.
Proceed cautiously once the earthquake has stopped. Avoid roads, bridges, or ramps that might have been damaged by the earthquake.
**If you are trapped under debris:**
Do not light a match.
Tap on a pipe or wall so rescuers can locate you. Use a whistle if one is available. Shout only as a last resort. Shouting can cause you to inhale dangerous amounts of dust.
**After an earthquake has occurred:**
It is not uncommon for aftershocks to occur. Know what to do if you are impacted by an aftershock – Drop, Cover and Hold On.
Stay safe by avoiding damaged buildings, fallen and damaged trees and powerlines, blocked or damaged roads and other hazards.
When cleaning, protect your health and safety. Wear strong boots, gloves and protective clothing and wash your hands and clothes regularly.
Electricity, gas and water supplies may be disrupted. If your property has been damaged, have all utilities checked and tested by a licenced technician before you use them.

*Appendix B: Text translated with the prompt "Translate into Chinese".*

确保在地震期间如果你在室内，请遵循以下步骤：
趴下；找一个坚固的桌子或其他家具进行掩护；并抓紧，直到震动停止。
趴下，掩护（在坚固的桌子或家具下）并抓紧。
保持在室内，直到震动停止并确保安全后再离开。
避免靠近外墙、窗户、悬挂物、镜子、高大的家具、大型电器和装有重物或玻璃的厨房橱柜。
除非你确定门框是坚固的承重门框并且离你很近，否则不要使用门口。许多室内门口结构较轻，无法提供保护。
保持在室内，直到震动停止并确认安全后再离开。震动期间不要离开建筑物。研究表明，大多数受伤发生在建筑物内的人试图移动到其他地方或试图离开时。
注意电力可能会中断，或自动喷水灭火系统或火警可能会启动。
有发展障碍/认知障碍/智力障碍的人：
如果你在理解、记忆或学习方面有困难，请随身携带一份简单的待办事项清单和重要信息，并将其放入你的应急包中。提前练习你的计划。如果你使用辅助沟通工具，请将这些工具纳入你的计划。
留在那里。
远离建筑物、路灯和电线。
一旦到了空旷的地方，待在那儿，直到震动停止。最大危险存在于建筑物外部、出口处和外墙旁边。
如果你在地震期间正在开车，请遵循以下步骤：
尽快在安全情况下停车，并留在车内。
避免在建筑物、树木、高架桥和电线附近停车。 地震停止后，谨慎前行。避开可能因地震而损坏的道路、桥梁或斜坡。
如果你被困在废墟下：
确保在地震期间或之后被困在废墟下时遵循以下步骤：
不要点燃火柴。
敲击管道或墙壁，以便救援人员找到你。如果有哨子，请使用哨子。大声呼救应作为最后的手段，因为大喊可能会让你吸入大量灰尘。
地震发生后： 余震的发生并不罕见。
了解如何应对余震的影响：
趴下 掩护 抓紧
余震的发生并不罕见。
了解如何应对余震的影响——趴下、掩护并抓紧。
通过避开受损建筑物、倒下的树木和电线、受阻或损坏的道路以及其他危险来保持安全。
清理时，保护你的健康和安全。穿上坚固的靴子、手套和防护服，并定期清洗双手和衣物。
电力、天然气和供水可能会中断。如果你的财产受到损坏，请在使用之前由持证技术人员检查和测试所有设施。

*Appendix C: Reduced text translated with the prompt "Translate into Spanish".*

**Si estás dentro en el interior durante un terremoto:**
AGÁCHATE al suelo; CÚBRETE bajo una mesa resistente u otro mueble; y SUJÉTATE hasta que el movimiento sísmico se detenga.
Agáchate, cúbrete (bajo una mesa resistente o un mueble) y sujétate.
Permanecer en interiores hasta que el temblor se detenga y estés seguro de que es seguro salir.
Evita paredes exteriores, ventanas, objetos colgantes, espejos, muebles altos, electrodomésticos grandes y gabinetes de cocina con objetos pesados o de vidrio.

No uses una puerta a menos que sepas que es una puerta con soporte fuerte, de carga, y esté cerca de ti. Muchas puertas internas están construidas de forma ligera y no ofrecen protección.
Permanecer en el interior hasta que el temblor se detenga y sea seguro salir. No salgas de un edificio durante el temblor. Las investigaciones han demostrado que la mayoría de las lesiones ocurren cuando las personas dentro de los edificios intentan moverse a otro lugar dentro del edificio o intentan salir.
Ten en cuenta que la electricidad puede apagarse o que los sistemas de rociadores o alarmas de incendio pueden activarse.
**Personas con discapacidades del desarrollo/cognitivas/intelectuales:**
Si tienes dificultad para entender, recordar o aprender, mantén una lista simple de qué hacer y la información importante contigo y en tus *kits*. Practica tu plan de antemano. Si utilizas apoyos de comunicación aumentativa, inclúyelos en tu planificación.
**Si estás al aire libre durante un terremoto:**
Permanecer allí.
Aléjate de edificios, farolas y cables de servicios públicos.
Una vez que estés en un área abierta, quédate allí hasta que el temblor se detenga. El mayor peligro existe justo afuera de los edificios, en las salidas y junto a las paredes exteriores.
**Si estás en un vehículo en movimiento durante un terremoto:**
Detente tan rápido como sea seguro y quédate en el vehículo. Evita detenerte cerca o debajo de edificios, árboles, pasos elevados y cables de servicios públicos.
Continúa con precaución una vez que el terremoto haya terminado. Evita caminos, puentes o rampas que puedan haber sido dañados por el terremoto.
**Si estás atrapado bajo escombros:**
Asegúrate de hacer lo siguiente si te encuentras atrapado bajo escombros durante o después de un terremoto:
No enciendas un fósforo.
Golpea una tubería o pared para que los rescatistas puedan encontrarte. Utiliza un silbato si tienes uno. Grita solo como último recurso. Gritar puede hacer que inhales grandes cantidades de polvo peligroso.
**Después de que ocurra un terremoto:**
No es inusual que ocurran réplicas.
Saber qué hacer si te ves afectado por una réplica
Agáchate
Cúbrete
Sujétate
No es inusual que ocurran réplicas. Saber qué hacer si te ves afectado por una réplica: Agáchate, cúbrete y sujétate.
Mantente a salvo evitando edificios dañados, árboles caídos y dañados, líneas eléctricas, caminos bloqueados o dañados y otros peligros.
Al limpiar, protege tu salud y seguridad. Usa botas fuertes, guantes y ropa protectora, y lávate las manos y la ropa regularmente.
El suministro de electricidad, gas y agua puede estar interrumpido. Si tu propiedad ha sido dañada, haz que todos los servicios sean revisados y probados por un técnico autorizado antes de usarlos.

## *Appendix D: Reduced text translated into Chinese with the discourse-specific prompt*

**地震来袭时的安全提示**

- **立即趴下**，找掩护并抓紧，直到震动停止。（**趴下、遮挡、抓紧！**）
- 如果在室内，**不要靠近窗户、外墙或高大的家具**。远离吊挂物、镜子、大型电器和厨房橱柜等有重物或玻璃的地方。

- 除非你确定门框是坚固的承重结构，否则**不要站在门口**。许多室内门框较轻，无法提供保护。
- **不要在震动期间离开建筑物**。研究显示，大多数伤害发生在人们试图移动或离开建筑物时。保持冷静，待在安全的地方，直到震动停止。
- 注意可能**断电，喷水灭火系统或火警**可能启动。

**特别提示：发展障碍/认知障碍/智力障碍人群**

- 随身携带简单的安全步骤清单和重要信息，提前演练你的计划。使用辅助沟通工具的人，请确保在计划中包括这些工具。

**地震时如果在车辆中：**

- 立即停车并**留在车内**。避免停在建筑物、树木、高架桥或电线附近。
- 地震停止后，谨慎前行，避开可能损坏的道路和桥梁。

**如果被困在废墟下：**

- **不要点火**。
- 敲击管道或墙壁，发出声音以便救援人员找到你。**有哨子时使用哨子**，大喊只作为最后手段，以避免吸入粉尘。

**地震后的安全提示：**

- 注意**余震的可能性**，记住**趴下、遮挡、抓紧**。
- 远离受损的建筑物、电线和树木，避开可能受损的道路。
- 清理时，保护好自己，穿戴**坚固的靴子、手套和防护服**，并定期清洗双手和衣物。
- **水、电、煤气供应可能中断**。若房屋受损，请在使用前由专业技术人员检查和测试所有设施。

## *Appendix E: Reduced text translated into Spanish with the discourse-specific prompt*

**¿Qué hacer si te encuentras dentro de un edificio durante un terremoto?**
AGÁCHATE en el suelo; CÚBRETE bajo una mesa resistente u otro mueble robusto; y SUJÉTATE hasta que el temblor pare.
Agáchate, cúbrete (bajo una mesa resistente o un mueble robusto) y sujétate.
Permanece dentro del edificio hasta que el temblor pare y estés seguro de que es seguro salir.
Evita las paredes exteriores, ventanas, objetos colgantes, espejos, muebles altos, electrodomésticos grandes y armarios de cocina con objetos pesados o de vidrio.
No uses el marco de una puerta, a menos que sepas que es un marco resistente y de carga, y esté cerca de ti.
Muchos marcos de puertas interiores son construcciones ligeras y no ofrecen protección.
No salgas de un edificio durante el temblor.
Las investigaciones muestran que la mayoría de las lesiones ocurren cuando las personas dentro de los edificios intentan moverse a otro lugar o intentan salir del edificio.
Ten en cuenta que la electricidad puede cortarse, o pueden activarse los sistemas de aspersores o alarmas contra incendios.
**Personas con discapacidades de desarrollo, cognitivas o intelectuales:**
Si tienes dificultades para comprender, recordar o aprender, mantén una lista simple de lo que debes hacer y otra información importante contigo y en tus kits de emergencia.
Practica tu plan con antelación.
Si usas apoyos de comunicación aumentativa, inclúyelos en tu planificación.
Permanece en el lugar hasta que el temblor termine y sea seguro moverse.
**¿Qué hacer si te encuentras al aire libre durante un terremoto?**
Aléjate de edificios, farolas y cables de servicios públicos.
Una vez que estés en un lugar abierto, quédate allí hasta que el temblor pare.
El mayor peligro se encuentra justo fuera de los edificios, en las salidas y cerca de las paredes exteriores.

**¿Qué hacer si te encuentras en un vehículo en movimiento durante un terremoto?**
Detente tan rápido como lo permita la seguridad y permanece dentro del vehículo.
Evita detenerte cerca o debajo de edificios, árboles, pasos elevados y cables de servicios públicos.
Continúa con precaución una vez que el terremoto haya terminado.
Evita carreteras, puentes o rampas que puedan haber sido dañadas por el terremoto.
¿Qué hacer si quedas atrapado bajo escombros durante o después de un terremoto?
No enciendas fósforos.
Golpea una tubería o una pared para que los rescatadores puedan localizarte.
Usa un silbato si tienes uno a mano.
Solo grita como último recurso, ya que gritar puede hacer que inhales grandes cantidades de polvo.
Después de un terremoto:
No es raro que ocurran réplicas.
**¿Qué hacer si te afecta una réplica?**
Agáchate, cúbrete y sujétate.
Mantente seguro evitando edificios dañados, árboles y líneas eléctricas caídas, carreteras bloqueadas o dañadas, y otros peligros.
Cuando limpies, protege tu salud y seguridad.
Usa botas resistentes, guantes y ropa protectora, y lava tus manos y ropa con regularidad.
Los suministros de electricidad, gas y agua pueden estar interrumpidos. Si tu propiedad ha sido dañada, asegúrate de que todos los servicios sean revisados y probados por un técnico autorizado antes de utilizarlos.